# Representing Knowledge about Norms

Daniel Kayser and Farid Nouioua[1]

**Abstract** Norms are essential to extend inference: inferences based on norms are far richer than those based on logical implications. In the recent decades, much effort has been devoted to reason on a domain, once its norms are represented. How to extract and express those norms has received far less attention. Extraction is difficult: as the readers are supposed to know them, the norms of a domain are seldom made explicit. For one thing, extracting norms requires a language to represent them, and this is the topic of this paper. We apply this language to represent norms in the domain of driving, and show that it is adequate to reason on the causes of accidents, as described by car-crash reports.

## 1. INTRODUCTION

Norms are essential in our life. Our everyday behavior is guided by our knowledge of the *normal* outcomes of an action, and our attention is naturally driven towards what we perceive as *abnormal* in a given situation.

A.I. has realized the importance of norms at several levels: early systems, like frames [13] and scripts [15] led to the development, in the late '70s and '80s, of non-monotonic reasoning systems with more clearly stated formal properties (e.g. [1, 3]). Their goal is to extend the set of conclusions beyond what is derivable on the strict basis of logical inference.

As the notion of norm is used in several contexts, we should make clear that we mean here norms that rule commonsense reasoning, but not norms used for legal reasoning, which are also a subject of growing interest in AI [2, 4, 6].

Little attention has been paid to the way to extract and express the norms of a given domain. Today, large amounts of texts concerning many domains are available for the computer. Extracting the norms from the texts is however a difficult problem: as the norms of a domain are generally supposed to be known by any reader, they are seldom made explicit.

Anyway, extracting them presupposes the existence of a language to represent them. The adequacy of a representation language must be evaluated. Showing that a reasoning system using the language is able to detect the same anomalies as those detected by a human reader is a good criterion.

An anticipated consequence of this work is to enrich the traditional (truth-based) approach to natural-language semantics. Inferences based on norms are indeed far richer than those based on logical implications. Consider e.g. the text:

*The car before me braked suddenly.*

Inferences based on truth contain statements like: "there exists a time $t$ and a car $c$ such that $c$ was before me at $t$, and $c$ braked at $t$." Inferences based on norms add, among many others: "both $c$ and me were driving in the same direction, with no other vehicle in between. I had to brake in order to avoid an accident."

Section 2 describes the domain, section 3 discusses the basic issues, section 4 defines the main features of the language, section 5 shows how it is used to detect anomalies, section 6 provides an example, section 7 describes the remaining steps of the project and opens perspectives of this work.

## 2. THE DOMAIN

We have selected the domain of driving, for the following reasons:

- The number of norms is presumably neither too large nor too small; they are not limited to those listed in the highway code,

- We have an easy access to an unlimited amount of short texts: every insurance company receives daily car-crash reports containing at most 5 lines describing the circumstances of an accident,

- Each report implies a number of facts, which are not logically entailed, requiring an abundant use of norm-based reasoning,

- The text reveals, both explicitly and implicitly, several anomalies, including of course the accident itself. The readers generally understand one of them as being "basic", and see it as the cause of the accident; the other anomalies are derived from it.

Designing a computer program that, for each text, discovers the same "basic" anomaly as human readers do, is an ambitious AI objective that nonetheless seems realistic.

The difficulty of the task consists in describing the domain with a relatively small number of predicates, while maintaining the possibility to discriminate among cases that look rather similar, but nonetheless call to mind distinct anomalies.

## 3. BASIC ISSUES

A car-crash report, as all texts, is a structure over propositions. Each proposition describes something about a continuously evolving world. Should the representation be based on discrete notions, reflecting the structure of the text, or on continuous ones, closer to physical reality? This dilemma is akin to naive vs. scientific physics. The choice of scientific physics is more comfortable: we can use what we have been taught about speed, acceleration, collisions, but here it is not only computationally expensive, but representationally inadequate. As a matter of fact, a physical model of the accident would require fixing the value of a number of parameters, which are neither present in, nor derivable from, the text. Even worse, the texts are not written by experts in mechanics, but by drivers likely to share the misconceptions about force and energy that are common in the population [10]: what makes sense for them may not be translatable in terms of scientific physics.

We therefore use a "naive" approach. The propositions are modeled by discrete states connected by a temporal relationship. A state is characterized by a set of literals that reflects static as well as dynamic properties; two states are distinct as soon as one of their literals has different truth-values.

[1] LIPN, UMR 7030 du CNRS, Institut Galilée, Université Paris-Nord, F-93430 Villetaneuse. email : {dk,nouiouaf@lipn.univ-paris13.fr}

A second dilemma concerns the nature of the relation between such states. The action described takes place, of course, in a linear time, but the writer of the text often spends much space to justify his/her behavior in terms of attempts to achieve desirable goals or to avoid undesirable ones. This is easily understandable, because the texts are intended to establish responsibilities in the accident. Now the states described as goals are not parts of the unfolding of the action. If we limit ourselves to the representation of actual events, we miss much of the substance of the text. On the other hand, creating a (sequence of) state(s) for every future that one of the protagonists is likely to have envisioned (like e.g. McDermott's chronicles [12]) would increase, without necessity in our case, the number of issues to be solved.

**Hypothesis**: the detection of anomalies requires only the representation of states-of-affairs presented as having really occurred; the potential or counterfactual states play an implicit role to determine the expectations of the protagonists, their reasons for acting or not acting, but need no explicit representation.
**Corollary**: the states can be represented by natural numbers.

The author focuses on anomalies, i.e. events which are considered as abnormal with respect to the 'normal' unfolding of events known by the reader. This phenomenon has motivated the notion of script [15], but modeling it remains a difficult problem. For instance, the number of scenes of a script, i.e. the granularity of an adequate representation, is often the result of understanding the text, not a prerequisite to understand it. Therefore the number of states keeps changing during the analysis.

# 4. THE LANGUAGE

To be able to quantify over variables representing names of predicates, and thus get a better factorization of the rules governing the domain, we made the choice of a first-order reified language.

Consequently, most of the statements will be under the form:

*Holds (P, X, t)*[2]

to state that *P* is true of *X* at state *t*.

However, not all statements are state-dependent; for instance, to represent the fact that whenever an effect F is observed, it is normally believed that an event V occurred, we use:

*Potentially_caused_by (F, V)* (abbr. *Pcb (F, V)*)

A last kind of statement concerns modalities. Instead of making use of modal connectives, we represent for instance the (moral) necessity for agent *X* to get effect *F* at state *t* by:

*Must (F, X, t)*

Finally, inference rules are either strict (no exception is likely to occur in the framework of our texts), or defeasible. Strict rules are rendered by material implications, and defeasible ones by Reiter's normal or semi-normal defaults [14]. The latter are needed as the semi-monotonicity of normal theories forbids them to cope with priorities among defaults [5].

*A : B* abbreviates the normal default: $\frac{A \,:\, B}{B}$

*A : B [C]*, the semi-normal default: $\frac{A \,:\, B \wedge C}{B}$

We now discuss with more details some of the predicates.

## 4.1. State-dependent predications

Norms, and hence anomalies, concern heterogeneous concepts of the domain. To reduce the complexity, we divide the problem into parts: issues related to the speed of vehicles in a file can be safely isolated from, say, considerations about priorities at crossroads. The predicates are therefore partitioned into layers. Layers are partially ordered from 'outside' (the expected result of a parser) to 'inside' (a dozen or so of state-dependent predicates constituting the *kernel*). The right-hand side of each inference rule contains predicates that are not more "external" than those of its left-hand side, thus favoring the convergence of the reasoning towards the predicates of the kernel.

The predicates of the kernel are chosen in such a way that every anomaly, once translated at their level, is still recognizable as an anomaly; naturally, many details are lost during the translation, and the explanation of the cause of the accident in terms of these predicates looks awkward.

## 4.2. State-independent predications

State-independent predicates describe the nature of predicates and their mutual relations. Reification allows providing some predicates with types, without appealing to second order. *Action* (*P*) is true iff *P* is the name of an action that an agent may perform voluntarily in order to achieve an effect. For instance, braking, turning the steering wheel are actions. Driving slowly or turning right are not actions, but effects of these actions. In our context, the number of actions is very limited.

*Event* (*P*) is true iff *P* is the name of an event. From the point of view of an agent, everything that happens independently of his/her will is an event; for instance, the moves of other agents, the outbreak of an obstacle in his/her visual field are events.

Actions and events are the only sources responsible of state changes, i.e. in the fact the truth-value of some literals, called effects, changes. We write *Effect* (*P*).

Effects caused by events may be undesired. The agent therefore may want to maintain the current state, in order to avoid such effects. For that purpose, a special action is at his/her disposal: for every effect *F*, *Combine* (*Keep_State*, *F*) is an action.

Among the effects resulting from action and events, some are persistent, unless another action or event causes them to change:

*Persistent (P) ∧ Holds (P, X, t) : Holds (P, X, t+1)*

This default expresses a forward persistence. Some effects are also backward persistent, but this is far less common. Therefore, backward persistence is expressed on a case-by-case basis.

To reason efficiently on causes and effects requires knowing how the different predicates involved are interrelated, independently of their occurrence in a temporal framework.

The relation *Incompatible* (*F, F'*) expresses that effects *F* and *F'* cannot be simultaneously true in any state. In particular, for all *F* we have: *Incompatible (F, Neg (F))*.

Causality is an extremely delicate issue, but we cannot escape it, as it plays a central role in the detection of anomalies [11]. What is needed here however, is not a relation Cause (*Act, F*) whatever this

---
[2] The arity of *Holds* restricts this notation to unary predicates. Where more arguments are required, we use a binary function *Combine*. For instance, to represent that the literal *Q (X, Y)* is true at state *t*, we write: *Holds (Combine (Q, Y), X, t)*. When needed, a function *Neg* applies on predicate names and we have: *Holds (Neg (P), X, t) ↔ ¬Holds (P, X, t)*.

may mean but, perhaps more simply, the expression of a belief, written *Pcb* (*F*, *P*) [for *Potentially caused by* as said earlier], where *P* is either an action or an event; this literal does not reflect a belief of unicity: many other causes may have produced *F*, and the goal is not to collect every factor yielding *P*, but only the ones that come straight to the mind of a standard reader. The limited extent of the domain keeps the number of such relations rather small, and we postulate that:

**Hypothesis**: At most one voluntary action *Act* satisfies *Pcb* (*F*, *Act*) [exception: if *F* is persistent, we also have *Pcb* (*F*, *Combine* (*Keep_State*, *F*))].

For an action *Act* to reach an effect *F* which it is known to be a potential cause of, the agent must perform it under adequate circumstances. This is the well-known *qualification* problem [8]. Here too, we neither want to, nor can, list these circumstances. Whenever some predicates *P* are likely to play a role to enable / prevent the success of an action, we write *Precond_Action* (*F*, *P*) for "*P* must be true for the (supposedly unique) action able to yield effect *F* to succeed" and *Precond_Av_Event* (*F*, *P*) for "*P* must be true for an agent to succeed in avoiding the effect *F* of an event".

## 4.3. Modalities

Modalities are central for detecting (basic and derived) anomalies. They are also helpful to reason on other elements of the language. Their analysis in terms of a kripkean semantics requires several types of accessibility relation between states. As this analysis does not shed more light on the problems discussed here, we omit it.

Modalities say something about forthcoming states, i.e. their being true at state *t* generally entails the truth of some (modal or non-modal) statements at state *t+1*.

### 4.3.1. Basic modalities

Norms and anomalies are related to what an agent must do. The literal *Must* (*F*, *X*, *t*) is true iff in state *t*, agent *X* has to get effect *F*. As agents are expected to comply with their duties, we have:

$$Must\ (F, X, t) : Holds\ (F, X, t+1).$$

The modality *Able_To* (*F*, *X*, *t*) expresses that in state *t*, agent *X* can do an action having the effect *F*. This literal is true even if the action eventually fails, as long as *X* cannot know that beforehand, i.e. has no excuse for not undertaking such an action.

### 4.3.2. Basic and derived anomalies

Basic anomalies come under two forms. The first one arises whenever an agent *X* must reach some effect *F* in state *t* and has the ability to reach it (in the sense of the above modality); however, at state *t+1* an effect *F'* incompatible with *F* holds true:

$$Must\ (F, X, t) \wedge Able\_To\ (F, X, t) \wedge Holds\ (F', X, t+1) \wedge$$
$$Incompatible\ (F, F') \rightarrow Anomaly$$

The second form corresponds to cases where a 'disruptive factor' (see §5.) exists for the agent:

$$Holds\ (Combine\ (Disruptive\_Factor, C), X, t) \rightarrow Anomaly$$

Derived anomalies correspond to situations where an agent did not fulfill his/her duties because s/he was not in position to comply:

$$Must\ (F, X, t) \wedge \neg Able\_To\ (F, X, t) \wedge Holds\ (F', X, t+1) \wedge$$
$$Incompatible\ (F, F') \rightarrow Derived\_Anomaly$$

### 4.3.3. Definition of the modality Able_To

The crucial point for the detection of anomalies consists in assessing whether an agent is in position to avoid a transition yielding an undesired state. More information about the features of the actions and events is generally needed to decide whether the modality *Able_To* holds.

The literal *Predictable* (*V*, *X*, *t*) expresses a property of event *V* that can be state-independent (e.g. icy patches could always be considered as unpredictable causes of loss of control); it can also be inferred in specific situations by means of appropriate rules (e.g. if *X* is an obstacle for *Y*, and *X* is not under control, then *X* is unpredictable for *Y*).

An event *V* is said to be *controllable* by agent *X* at state *t* iff: either *V* does not occur at time *t*, or it was predictable and *X* is in position at state *t* to satisfy the precondition of its avoidance.

$$Event\ (V) \wedge (\neg Holds\ (V, X, t) \vee (Predictable\ (V, X, t) \wedge$$
$$(Precond\_Av\_Event\ (F, P) \rightarrow Holds\ (P, X, t))) \leftrightarrow$$
$$Controllable\ (V, X, t).$$

An agent may undertake an action *Act* without knowing whether it will succeed. The predicate *Available* is meant to express that, as far as the agent knows, *Act* satisfies all its preconditions. The literal *Available* (*Act*, *F*, *X*, *t*) is thus true iff if at state *t*, agent *X* decides to execute *Act* with the belief that effect *F* will obtain. A default assumption is that every action is available:

$$Pcb\ (F, Act) \wedge Action\ (Act) : Available\ (Act, F, X, t).$$

This assumption forces to enumerate the situations where an action is not available. Several cases of unavailability are considered:

- the presence of 'technical problems":

$$Holds\ (Combine\ (Tech\_Pb, Act), X, t) \wedge Holds\ (Act, X, t) \wedge Pcb\ (F, Act) \rightarrow \neg Available\ (Act, F, X, t)$$

- the precondition of the action being not satisfied:

$$Action\ (Act) \wedge Precond\_Action\ (F, P) \wedge \neg Holds\ (P, X, t) \rightarrow$$
$$\neg Available\ (Act, F, X, t)$$

- a keep-state action is available except if an uncontrollable event leads to a state where *F'* holds, and *F'* is incompatible with *F*:

$$(\exists F', V)\ (Pcb\ (F', V)\ \wedge Event\ (V)\ \wedge \neg Controllable\ (V, X, t) \wedge$$
$$Incompatible\ (F, F')) \leftrightarrow \neg Available\ (Combine\ (Keep\_State, F), F, X, t)$$

- the loss of control of a vehicle makes every action of its driver obviously unavailable:

$$\neg Holds\ (Control, X, t) \wedge Pcb\ (Act, F) \rightarrow \neg Available\ (Act, F, X, t)$$

These predicates delimit the states where an agent is *Able_To* undertake an action. Intuitively, agent *X* is able to reach effect *F* at state *t* iff there exists an action *Act* that is a potential cause for *F* and is available for *X* at *t*:

$$Able\_To\ (F, X, t) \leftrightarrow (\exists Act)\ (Action\ (Act) \wedge Available\ (Act, F, X, t)$$
$$\wedge Pcb\ (F, Act))$$

## 5. DETECTING ANOMALIES

Unpredictable events causing a loss of control are among what we called disruptive factors:

*Holds (Combine (Cause_No_Control, C), X, t) ∧ ¬Predictable (Combine (Cause_No_Control, C), X, t) → Holds (Combine (Disruptive_Factor, C), X, t)*

So-called "technical problems" are treated that way. They are unpredictable causes of loss of control, and this is enough (see §4.3.2.) to assign to them the responsibility of a basic anomaly. Unpredictable obstacles other than vehicles are disruptive factors as well (in the case of vehicles, other factors are privileged):

*Holds (Combine (Obstacle, O), X, t) ∧ ¬Predictable (Combine (Obstacle, O), X, t) :*
*Holds (Combine (Disruptive_Factor, O), X, t) [¬Vehicle (O)]*

A derived anomaly (§4.3.2.) always occurs at the transition between two states *t* and *t*+1; the basic anomaly from which it derives often occurs at the preceding transition *t*-1, *t*. Therefore an abductive reasoning hypothesizes that the precondition of an action *Act* was not satisfied at state *t*, and attempts to find what went wrong at state *t*-1. For instance, if *Act* is a keep-state action, abduction works through two rules:

- The first one concerns the case of a predictable event producing an effect *F'* incompatible with the desired effect *F*. If the agent has control but is unable to reach *F*, the rule abducts that the precondition for avoiding the consequence of the event was not satisfied.

*Holds (Control, X, t) ∧ Must (F, X, t) ∧ ¬Able_To (F, X, t) ∧ Holds (V, X, t) ∧ Event (V) ∧ Pcb (V, F') ∧ Incompatible (F, F') ∧ Predictable (V, X, t) ∧ Precond_Av_Event (F', P)*
*→ ¬Holds (P, X, t)*

- The second rule propagates duties backwards: if at some state *t*, the agent must obtain effect *F* and a predictable event is known to produce effect *F'* incompatible with *F*, then the agent must satisfy at state *t*-1 the precondition avoiding the consequence of the event.

*Must (F, X, t) ∧ Holds (V, X, t) ∧ Event (V) ∧ Pcb (V, F') ∧ Incompatible (F, F') ∧ Predictable (V, X, t) ∧ Precond_Av_Event (F', P) → Must (P, X, t-1)*

Similar abductive rules are written for situations where the duty of the agent is not to counteract a predictable event, but to execute an action in order to get an effect *F*. If the agent has control, meets no technical problem, but does not get the effect, the rule concludes that the precondition was not satisfied:

*Holds (Control, X, t) ∧ Must (F, X, t) ∧ ¬Holds (F, X, t+1) ∧ Pcb (Act, F) ∧ Holds (Act, X, t) ∧ ¬Holds (Combine (Tech_Pb, Act), X, t) ∧ Precond_Action (F, P) → ¬Holds (P, X, t)*

Finally, the backpropagation of duties is expressed here by:

*Must (F, X, t) ∧ ¬Holds (F, X, t) ∧ Precond_Action (F, P) → Must (P, X, t-1).*

## 6. EXAMPLE

I was beginning to turn right when I saw Mr.L's car coming in the opposite direction and encroaching on my lane. As I was driving slowly, I stopped at once. Mr.L, who drove faster, was unable to act similarly, and rubbed his car all along on my front bumper.

I was not in position to catch sight of Mr.L earlier, because he was driving on his left (he was overtaking a parked vehicle) in a street masked by a hedge.

This text, the first report of our corpus, implies at least three states. For all texts, state 0 contains the default assumption that every vehicle is under control, and no vehicle is stopped:

*Holds (Control, X, 0), ¬Holds (Stop, X, 0) (X ∈ {A, B})*

Predicates *Control* and *Neg (Stop)* are declared as *Persistent*. The above literals thus remain true in the other states, unless proof of the contrary. The first state explicitly mentioned in the text, state 1, contains the fact that *A* is turning right[3]:

*Holds (Combine (Turn, right), A, 1)*

At the same state 1, the text says that A and B drive in opposite directions, and that B was partly on A's normal lane, from which one derives that B was not completely on its normal lane (we learn later that it is because B was overtaking a parked vehicle):

*¬Holds (Combine (Same_Way, A), B, 1), Holds (Combine (Same_Lane, A), B, 1), ¬Holds (On_Normal_Lane, B, 1), Holds (Is_Overtaking, B, 1)*

*Same_Lane* has not the expected meaning that both vehicles are entirely on the same lane; it is satisfied as soon as at least part of the vehicles is located on the same lane — this is the important fact to track anomalies —. According to the next sentence, at state 1, vehicle *A* drove fairly slowly, and this was not the case of *B*.

*Holds (Drive_Fairly_Slow, A, 1), ¬Holds (Drive_Fairly_Slow, B, 1)*

The text then presents a state 2, where *A* stopped, *B* did not stop, and there was a shock between *A* and *B*:

*Holds (Stop, A, 2), ¬Holds (Stop, B, 2),*
*Holds (Combine (Shock, A), B, 2)*

We have a rule that says that every vehicle must be on its normal lane, except if it is overtaking:

*¬Holds (Stop, A, t) : Must (On_Normal_Lane, A, t) [¬Holds (Is_Overtaking, A, t)].*

This default is blocked for B at state 1, since the author provides the necessary information, but it works for *A* at state 0 and yields *Must (On_Normal_Lane, A, 0)*. The default given in §4.3.1. uses this fact to conclude: *Holds (On_Normal_Lane, A, 1)*. No anomaly is detected for the moment. But another rule tells that when two vehicles share the same lane and in opposite ways, they must stop:

*Holds (Combine (Obstacle, X), Y, t) ∧ Holds (Combine (Same_Lane, X), Y, t) ∧ ¬Holds (Combine (Same_Way, X), Y, t) → Must (Stop, X, t) ∧ Must (Stop, Y, t),*

---
[3] Actually, the text mentions the « beginning » of the turn; this is a kind of rhetorical figure, and we neglect it.

where *Obstacle* is defined in a rather extensive way: if two vehicles crashed at state *t*, the default assumption is that they were obstacles for each other at state *t*-1. This is expressed by:

*Holds (Combine (Shock,X), Y, t) : Holds (Combine (Obstacle, X), Y, t-1) [Vehicle (Y)]*

We now get: *Must (Stop, A, 1)* and *Must (Stop, B, 1)*. As we have *Holds (Stop, A, 2)* and ¬*Holds (Stop, B, 2)*, we know that contrary to *B*, *A* complied with his duty. We have nearly all we need to find the anomaly, in the sense of §4.3.2. Whether it is a basic or a derived one depends on what *B* is *Able_To* do. The definition of *Able_To* (§4.3.3.), instantiated with *F = Stop X = B*, *t = 1*, shows that the answer depends on the existence of an action *Act*, available for *B*, and such that *Stop* is potentially caused by it. We do have *Pcb* (*Stop, Brake*), so the question is whether or not *Brake* is available for *B*. The precondition for *Brake* at *t* to reach *Stop* at *t+1* is that the agent drives fairly slowly, i.e. *Precond_Action* (*Stop, Drive_Fairly_Slow*). The second case of unavailability of §4.3.3. has all its premises satisfied, thus derives that *Brake* is not available for *B* at state 1 (in the special sense given here, that is: *Brake* will not achieve *Stop* whereas it is the action known to reach this goal). The uniqueness hypothesis (§4.2.) allows using the last formula of §4.3.3. to conclude ¬*Able_To* (*B, Stop*, 1), and by 4.3.2., we get the answer: *Derived_Anomaly*.

The basic anomaly remains to be found, and it will be found by abduction. The last rule of §5 with *F = Stop*, *X = B*, *t = 1*, *P = Drive_Fairly_Slow* concludes: *Must (Drive_Fairly_Slow, B, 0)*.

We know that *Pcb* (*Brake, Drive_Fairly_Slow*). Notice that it is not a violation of the uniqueness hypothesis: depending on the situation, the outcome of the action *Brake* can be a slow down or a stop, but either one of these goals is reached by only one action.

The default in §4.3.3. gives *Available* (*Brake, Drive_Fairly_Slow, B*, 0) whence *Able_To* (*Drive_Fairly_Slow, B*, 0). The first rule of §4.3.2. derives *Anomaly* and the reasoning stops.

To sum up, the basic anomaly, i.e. what the author of this report suggests as the cause of the accident, is that, at the beginning of the episode, vehicle *B* could brake and did not so. This is the reason why, in state 1, *B* could not stop, causing the accident in state 2.

## 7. CONCLUSION and PERSPECTIVES

Space limitation forbids us to present the architecture of the system in progress of implementation, and a detailed status of each of its modules. We are, as it were, digging a gallery from both ends: from the linguistic end, where we have adapted an existing tagger for French, and written a special-purpose parser, and from the logical end, where we have defined, above the kernel, two more layers: layer 2 copes with priorities, visibility, lanes, obstacles, various causes of loss of control; layer 3 deals with positions of vehicles to derive predicates of layer 2. We have also designed a set of around 50 predicate names that constitutes the language where the two ends of the gallery should meet.

We have analyzed manually 60 reports of our corpus and designed around 100 rules allowing to find the anomaly that human readers take as the reported cause of the accident. This result is not meaningful yet, because the rules were crafted after examination of the corpus; we will shortly collect many more reports from insurance companies to check on a wider corpus whether this result remains valid.

Except for 'time dilatation", i.e. the fact that during the reasoning, we have to insert states between those that result from the linguistic analysis of the text, the example presented in §6. shows more or less every difficulty encountered. The result of the parser inevitably comes with spurious analyses, but we are fairly confident that most of them will merge into the same logical form after a couple of inferences. Crude filters are being tested, in order to eliminate the parts of the report that are purely argumentative. The automatic identification of states, from linguistic (e.g. grammatical tenses [9], conjunctions) and extra-linguistic clues will be the next truly difficult issue to tackle.

An inference engine will handle the facts and rules. Although non-monotonic reasoning systems belong to intractable complexity classes, we are rather optimistic, as we have found, on a sample of reports including head-on crashes, refusals to yield way at an intersection, pulling out while being overtaken, etc., that a rather small amount of distinct state-independent predications was enough to cover a variety of cases. Moreover, the size of the Herbrand universe for this kind of applications is small; the predicates being stratified in layers, it should be easy to determine early in the process which defaults are blocked. Finally, completeness is not a crucial issue, since the reasoning is stopped as soon as the literal *Anomaly* is derived; simple heuristics [7] might therefore speed up the process, while keeping a reasonable rate of success.

If a domain of the size of the one explored here can be handled by a few hundreds rules, this opens the possibility to express the norms, and thus to enrich the power of inference engines, for many other domains of our everyday life.

## ACKNOWLEDGEMENTS

Many thanks to Françoise Gayral, François Lévy, Catherine Recanati, for their helpful suggestions on earlier drafts of this paper. This work is part of the 02 MDU 542 project of the franco-algerian cooperation CNRS-CMEP.